\documentclass[10pt,twocolumn,letterpaper]{article}

\usepackage[pagenumbers]{wacv} 

\usepackage[dvipsnames]{xcolor}

\newcommand{\boost}{SimpleBoost}
\newcommand{\vect}[1]{{\bf #1}}

\usepackage{wrapfig}

\definecolor{wacvblue}{rgb}{0.21,0.49,0.74}
\definecolor{orange}{rgb}{1,0.5,0}
\usepackage[pagebackref,breaklinks,colorlinks,allcolors=wacvblue]{hyperref}
\usepackage[table]{xcolor}
\title{EfficientDepth: A Fast and Detail-Preserving \\Monocular Depth Estimation Model}

\author{Andrii Litvynchuk$^{1}$ \hspace{0.2in} 
Ivan Livinsky$^{1}$ \hspace{0.2in}
Anand Ravi$^{1}$ \hspace{0.2in}
Nima Khademi Kalantari$^{2}$ \hspace{0.2in} 
Andrii Tsarov$^{1}$\\
$^1$Leia Inc. \hspace{0.2in}
$^2$Texas A\&M University\\
{\tt\small \{andrii.litvynchuk, ivan.livinsky, anand.ravi, andrii.tsarov\}@leiainc.com \hspace{0.2in} nimak@tamu.edu}
}

\DeclareMathOperator{\argmax}{argmax}
\newcommand{\LPIPS}{\mathrm{LPIPS}}

\begin{document}
\maketitle
\begin{abstract}
Monocular depth estimation (MDE) plays a pivotal role in various computer vision applications, such as robotics, augmented reality, and autonomous driving. Despite recent advancements, existing methods often fail to meet key requirements for 3D reconstruction and view synthesis, including geometric consistency, fine details, robustness to real-world challenges like reflective surfaces, and efficiency for edge devices. To address these challenges, we introduce a novel MDE system, called EfficientDepth, which combines a transformer architecture with a lightweight convolutional decoder, as well as a bimodal density head that allows the network to estimate detailed depth maps. We train our model on a combination of labeled synthetic and real images, as well as pseudo-labeled real images, generated using a high-performing MDE method. Furthermore, we employ a multi-stage optimization strategy to improve training efficiency and produce models that emphasize geometric consistency and fine detail. Finally, in addition to commonly used objectives, we introduce a loss function based on LPIPS to encourage the network to produce detailed depth maps. Experimental results demonstrate that EfficientDepth achieves performance comparable to or better than existing state-of-the-art models, with significantly reduced computational resources.



\end{abstract}    
\section{Introduction}
\label{sec:intro}
Monocular depth estimation (MDE) is a fundamental task in computer vision with a broad range of applications, including autonomous driving, virtual/augmented reality, and robotics. Recent years have seen significant advancements in MDE, largely driven by the availability of large-scale datasets that enable the training of deep networks to infer depth from single images~\cite{DA1, DA2, DepthPro}.

Despite these advancements, existing approaches often fall short of meeting certain requirements crucial for 3D reconstruction and view synthesis applications: \textbf{1)} the depth estimates must be geometrically consistent and highly detailed with pixel-perfect boundaries, \textbf{2)} the method should perform well on real images and handle challenging cases such as reflective surfaces, and \textbf{3)} the network must be highly efficient to ensure reliable operation on edge devices.

To address these needs, we carefully design our network architecture to be both powerful and capable of producing highly detailed depth maps across diverse scenarios, while also being efficient. Specifically, we use a combination of a transformer architecture and a lightweight convolutional decoder. Additionally, inspired by the approach of Tosi et al.~\cite{SMD}, we introduce a bimodal density head to enhance the network's ability to produce sharp boundaries.

Similar to existing methods~\cite{DA1, DA2}, we train our network on a collection of labeled synthetic and real images, as well as pseudo-labeled real images. However, departing from current methods, we create our pseudo labels using a high-performing monocular depth estimation method~\cite{DA1} and propose a simple yet effective boosting strategy to enhance the results. Additionally, we ensure that reflective surfaces, for which correct labels only exist in synthetic data, are included in our synthetic examples. Moreover, we perform optimization in multiple stages to accelerate the training process and produce models that emphasize specific strengths, such as geometric consistency and fine detail. Finally, in addition to common objectives, such as scale- and shift-invariant absolute error~\cite{Midas} and edge-matching loss~\cite{EdgeLoss2}, we propose a loss based on LPIPS~\cite{LPIPS} to enhance detail. We demonstrate that our approach outperforms existing methods, while being more efficient.

\section{Related work}
\label{sec:related}
MDE has evolved from simple perceptual models to sophisticated, data-driven deep learning approaches. Early methods~\cite{Hoiem, SiftFlow} relied on geometric techniques from classical computer vision and manually crafted features. Later models shifted to Deep Learning, using various architectures and large volumes of training data.

An early convolutional MDE model~\cite{Eigen} featured a relatively simple architecture and generated metric depth predictions. However, it was limited to narrow domains since the training data consisted of only two early datasets.

Midas~\cite{Midair} introduced a scale- and shift-invariant loss that enabled training on a diverse collection of mixed labeled datasets, which led to significantly improved accuracy. In our work, we apply this scale- and shift-invariant loss to train our model across various distinct datasets, building on the advancements made by Midas. ZoeDepth~\cite{ZoeDepth} enhanced absolute depth estimation accuracy by applying fine-tuning with metric depth data, which helped the model generalize better in real-world scenarios. Other approaches, like AdaBins~\cite{AdaBins} and BinsFormer~\cite{BinsFormer}, framed MDE as a regression-classification task, where the entire depth range is discretized into bins, allowing for more structured depth estimation across varying ranges.

Recent generative approaches, such as DiffusionDepth~\cite{DiffusionDepth} and Marigold~\cite{Marigold}, leverage the extensive visual knowledge embedded in modern generative models. For example, Marigold is derived from Stable Diffusion and fine-tuned with synthetic data, achieving high accuracy through a tailored fine-tuning protocol for depth estimation.

The DepthAnything V1 model~\cite{DA1} introduced a student-teacher approach that demonstrated the value of large-scale unlabeled data. Rather than new technical modules, it achieved significant progress by scaling the dataset, using an automated engine to collect and annotate large volumes of data. We also use pseudolabeled data in our model, following the core ideas from \cite{DA1}. Building on this, DepthAnything V2~\cite{DA2} provides versatile depth estimation models across multiple sizes (from small to giant), allowing adaptation to varying computational requirements.


Finally, Depth Pro~\cite{DepthPro}, a concurrent MDE model, synthesizes high-resolution depth maps with fine details. Notably, its predictions are metric and do not require camera metadata, such as intrinsics, making it versatile for applications lacking such data.
\section{EfficientDepth}

The primary objective of our work is to design an MDE model with key properties necessary for effective use in 3D reconstruction and view synthesis applications, including geometric consistency (i.e., ensuring accurate relative depth prediction and correct object positioning within the scene), fine detail, robustness in handling real images with reflective surfaces, and efficiency for edge devices. 

Our approach takes a single image as input and, similar to existing methods~\cite{Midas,DA1,DA2}, predicts the inverse depth of the scene, $\hat{d}$, a scaled disparity. This representation is particularly suited for multi-dataset training where labels are provided in various forms, including absolute depth, depth up to a scale, and disparity. Directly estimating depth requires inverting the disparity provided by certain datasets, which can be complex, as disparity depends on convergence (focus) and may vary in sign when the convergence point lies on an object within the scene. In contrast, depth, which is always positive, can be easily converted to a randomly scaled disparity using $\frac{1}{\text{depth}}$.

In the following sections, we discuss our approach by explaining the network architecture (Sec.\ref{ssec:architecture}), dataset (Sec.\ref{ssec:dataset}), training strategy (Sec.~\ref{ssec:training}), and training objectives (Sec.~\ref{ssec:objectives}).


\subsection{Network Architecture}
\label{ssec:architecture}

Our main goal is to design a network that is both efficient and capable of producing detailed and geometrically consistent depth maps. Although the fully transformer-based architectures in previous techniques~\cite{Midas} are powerful, they are not designed with efficiency in mind. In our system, we employ MiT-B5~\cite{SegFormer}, a lightweight transformer network, as the encoder and use a simple UNet~\cite{UNet} as the decoder. The key idea is that once the transformer extracts powerful features by capturing long-range dependencies, the UNet can decode these features and produce the final results.

Since our network is relatively small, we observed that in some cases it struggles to estimate sharp discontinuities. To address this, inspired by Tosi et al.~\cite{SMD}, we propose adding a final layer that enables the network to estimate a distribution with two modes over disparities, represented as a bimodal Laplacian mixture. Specifically, given a set of features estimated by the decoder at each pixel (five channels), we use a multilayer perceptron (MLP) with ReLU activations to estimate a five-dimensional vector $(\pi, \mu_1, b_1, \mu_2, b_2)$, representing the parameters of a bimodal distribution:

\vspace{-0.15in}
\begin{equation}
  p(d) = \frac{\pi}{2b_1} e^{-\frac{|d-\mu_1|}{b_1}}
+ \frac{1-\pi}{2b_2} e^{-\frac{|d-\mu_2|}{b_2}}.  
\end{equation}
\vspace{-0.15in}

\noindent The final disparity is then estimated as follows:

\vspace{-0.15in}
\begin{equation}
    d^* = \underset{d\in\{\mu_1, \mu_2\} }{\argmax}p(d).
\end{equation}
\vspace{-0.15in}

Here, our network estimates two distinct disparity maps, along with a probability for selecting one at each pixel. This provides the network with the flexibility to estimate smoother disparity maps, as the sharp boundaries can be obtained by making a hard decision when selecting one of the two modes of the distribution.

\begin{table}
  \centering

\caption{The datasets we used for training and testing in our work}
\vspace{-0.1in}
  \begin{tabular}{lccc}
    \toprule
    Title & Amount  & Type & Split\\ 
    \midrule
    DIML\cite{DIML} & 188k & Real & train\\
    Stereo Datasets & 51k & Real-S & train\\
    MegaDepth\cite{MegaDepth} & 88k & Real & train\\
    SceneFlow\cite{SceneFlow} & 33k & Synth & train\\
    TartanAir\cite{TartanAir} & 307k & Synth & train\\
    IRS\cite{IRS} & 103k & Synth & train\\
    MidAir\cite{Midair} & 124k & Synth & train\\
    HyperSim\cite{Hypersim} & 72k & Synth & train\\
    Pano3d\cite{Pano3d} & 58k & Synth & train\\
    Omniverse\cite{Omniverse} & 68k & Synth & train\\
    ClearGrasp\cite{Cleargrasp} & 45k & Synth & train\\
    Niklaus\cite{Niklaus} & 76k & Synth & train\\
    Eden\cite{Eden} & 184k & Synth & train\\
    Sports & 126k & Synth & train\\
    FaceSynthesis\cite{FaceSynthesis} & 80k & Pseudo & train\\
    FaceReal & 35k & Pseudo & train\\
    COCO\cite{COCO} & 247k & Pseudo & train\\
    Art & 248k & Pseudo & train\\
    WSVD\cite{WSVD} & 312k & Pseudo & train\\    
    Google Landmarks\cite{Google}& 719k & Pseudo & train\\
    ImageNet-21K\cite{ImageNet} & 2m & Pseudo & train\\
    OpenImages\cite{OpenImages} & 1m & Pseudo & train\\
    SA-1B\cite{SA1B} & 1.5m & Pseudo & train\\
    \midrule
    KITTI\cite{KITTI} & 161 & Real & test\\
    ETH3D\cite{ETH3D}  & 454 & Real & test\\
    DIW\cite{DIW} & 74k & Real & test\\
    NYUv2\cite{NYUDV2} & 654 & Real & test\\
    TUM\cite{TUM} & 1.8k  & Real & test\\
    Sintel\cite{Sintel}  & 1k  & Synth & test\\
    \bottomrule
  \end{tabular}
\vspace{-0.2in}
  \label{tab:datasets}
\end{table}

\subsection{Dataset}
\label{ssec:dataset}

We use 23 datasets for training and a separate set of 6 datasets for testing our approach, as summarized in Table~\ref{tab:datasets}. For training, we use four types of datasets, described below:

\paragraph{Real:} These datasets contain real images where depth data is obtained using LiDAR, COLMAP, or partially labeled. Real data is essential for enabling the network to adapt to real images; however, it is generally sparse and, in some cases, noisy or inaccurate~\cite{DA2}. Due to these issues, we use only one real single-image dataset and achieve the majority of adaptation through pseudo-labeled images.

\paragraph{Real-S:} These datasets consist of real \emph{stereo} images for which we estimate the corresponding disparity using CREStereo~\cite{Crestereo}. This approach produces detailed disparity with reasonable accuracy, making these datasets valuable for training. We use a combination of Flickr1024~\cite{Flickr1024} and HoloPix~\cite{Holopix} as our real stereo data. Note that only the left image of the stereo pair is used for training.

\paragraph{Synth:} These datasets contain synthetically generated images with detailed and geometrically accurate depth information. Synthetic data is crucial for learning to predict fine depth details, though relying solely on synthetic data may lead to poor generalization on real-world examples. The \textit{Sports} dataset, specifically, is a collection of football and basketball images sourced from video games (FIFA and NBA 2K) with ground-truth depth.

\paragraph{Pseudo:} These are a collection of images without ground truth depth, for which we estimate pseudo labels using the strategy, discussed in Sec.~\ref{ssec:pseudo}. Here, \textit{FaceReal} is a combination of four face detection datasets: FDDB~\cite{FDDB}, SoF~\cite{Sof}, UFDD~\cite{UFDD}, and WIDER FACE~\cite{WIDER_FACE}, while \textit{Art} consists of cartoon-style and animation scenes sourced from stock image collections. We plan to make this dataset publicly available to promote fair usage and reproducibility.

\begin{figure}
  \centering
  \includegraphics[width=\linewidth]{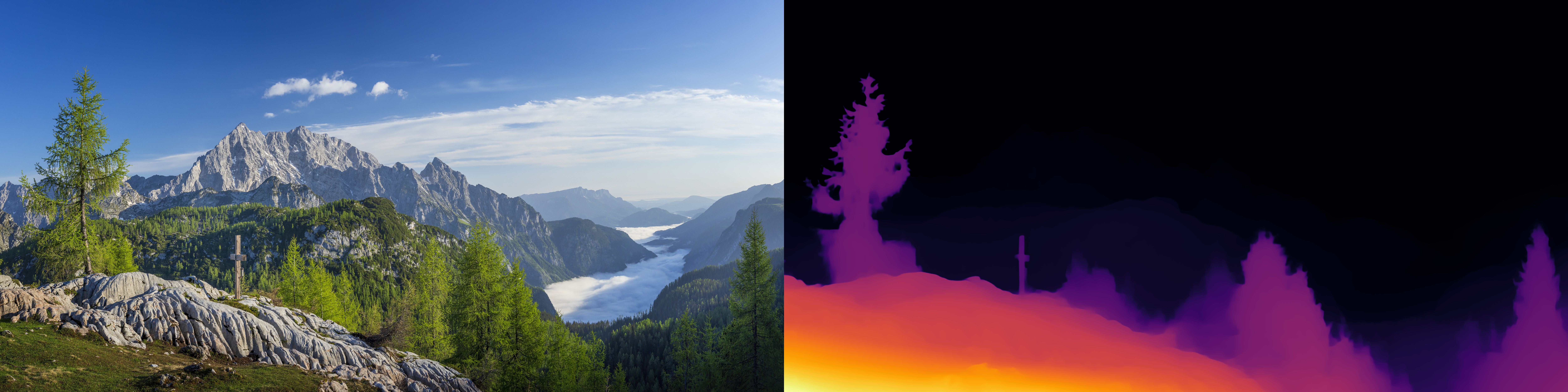}
  \includegraphics[width=\linewidth]{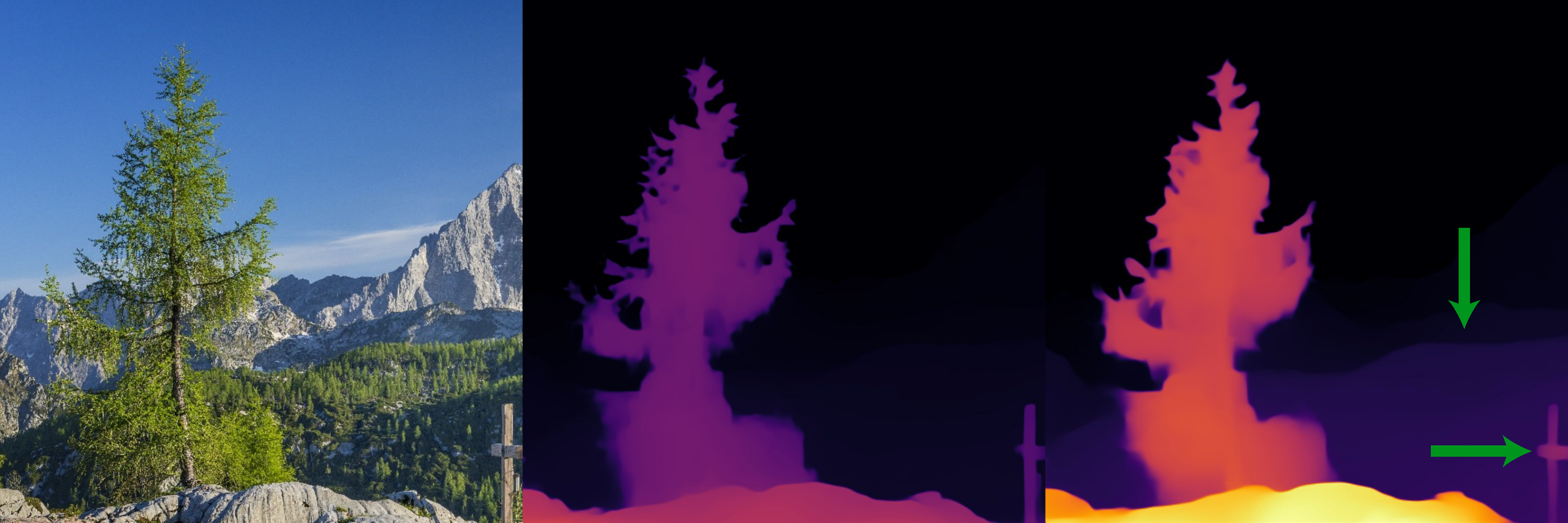}
  \hfill
  \vspace{-0.2in}
  \caption{\textit{First row}: Original image (1024$\times$2048), Depth Anything V1 Large (DA1L) applied at 2048 resolution. \textit{Second row}: The 640$\times$640 patches extracted from each of the images in the upper row, and DA1L applied on this patch at 518 resolution. Running DA1L on patch yields better background handling (more details), and better consistency (e.g. cross).} 
  \label{fig:patches}
  \vspace{-0.2in}
\end{figure}

\subsubsection{Enhanced Pseudo Labeling with \boost}
\label{ssec:pseudo}

Most recent MDE methods~\cite{DA1,DA2} utilize pseudo labeling to improve network performance on real-world examples. Specifically, they first train their networks on synthetic and/or labeled real images, then use the trained network to produce pseudo labels for real, unlabeled images. These pseudo labeled images are subsequently used to further fine-tune the network.

We adopt a similar strategy, however, unlike these methods, we propose using an existing MDE model to generate our pseudo labels. Since these models are fully trained and already adapted to real images, we found them to be a better choice than our model for this task. Among the methods we tested, Depth Anything V1~\cite{DA1} Large produced the most geometrically consistent depth maps, and thus we selected it to generate the pseudo labels.



Unfortunately, all existing methods, including Depth Anything V1 Large, perform optimally only at specific resolutions. Beyond this optimal range, they often yield results that lose geometrical consistency and contain fewer details at higher resolutions (see Fig. \ref{fig:patches}). This limitation can negatively impact the performance of our trained network, especially since a large portion of the images in our pseudo datasets are high-resolution.

To address this issue, we propose a simple and effective strategy, called \boost, which combines depth estimates from patches to create a coherent high-resolution depth map. Specifically, we found that Depth Anything V1 Large performs well at an image size close to $640 \times 640$ (see Fig.~\ref{fig:patches}). Thus, we propose to divide each high-resolution image into a grid of $640 \times 640$ patches with a $320$-pixel overlap, estimating depth for each patch individually as $D_{\text{v1}}(T_i(\vect{I}))$, where $T_i$ is a cropping operator that extracts the $i^{\text{th}}$ patch from the image $\vect{I}$.

While these patch-based depth estimates are generally detailed and high quality, they often exhibit inconsistent value ranges. To address this, we use the depth estimate from the full downsampled (using DepthAnything's native working size - 518) image as a reference and globally align each patch to its corresponding region through a scale and offset. We compute the optimal scale and offset for each patch as follows:

\vspace{-0.1in}
\begin{equation}
    s^*_i, o^*_i = \arg\min_{s_i, o_i} \Vert T_i(D_{\text{v1}}(\vect{I}\downarrow)\uparrow) -   (s_i D_{\text{v1}}(T_i(\vect{I})) + o_i)\Vert_2,
\end{equation}
\vspace{-0.1in}

\noindent where $\downarrow$ and $\uparrow$ denote the downsampling and upsampling operators, respectively. Since this is a quadratic objective, the solutions for $s_i$ and $o_i$ can be obtained in closed form.

With the optimal scale and offset for each patch, we can now obtain a set of globally aligned patches. To reconstruct the final high-resolution depth map, we linearly blend the patches in the overlapping regions to ensure smooth transitions. Note that the depth estimate from the full downsampled image is used only for alignment; the final depth map is constructed solely from the patch depth estimates.

Unlike the previous boosting method~\cite{Boosting}, our approach does not rely on a separate network to combine the patches, which mitigates potential issues with generalization. Our method generally yields similar or better results and is more efficient. You can find more details in the supplementary.

\subsection{Training}
\label{ssec:training}

We propose to perform training in three stages to reduce training time and provide different models with specific strengths. Below we describe each stage in detail:

\paragraph{Stage 1 - Main Training:} In this stage, we utilize all available data—both labeled and pseudo-labeled images. This is the longest and most crucial part of the training, as the network learns geometric consistency across scenes. Given its importance, we incorporate our entire dataset, which consists of approximately 8 million images (2 million labeled and 6 million pseudo-labeled). For this stage, we train for 16 epochs with a $320\times 320$ image size, a batch size of 500, and a learning rate of 0.0025.

\paragraph{Stage 2 - Resolution Adaptation:} In this stage, we reuse the same data from the main training stage, but increase the image size to $736\times 736$, training for only 2 epochs. This is our primary inference image size, and we found it important for the network to learn at this resolution, even if briefly. Although this stage could theoretically be merged with the main training stage, our experiments showed that doing so would yield almost identical results while increasing training time by a factor of 3. For this stage, we use a batch size of 70 and a learning rate of 0.00075. We also employ square root scaling for the learning rate, adjusting it based on batch size, which proved to be an effective approach.

\paragraph{Stage 3 - Fine Details Learning:} We use only synthetic data in this stage to focus on the highest-quality data without relying on imperfect pseudo-labels or real data, thereby ensuring the best possible detail in depth predictions. Limiting this stage to just four epochs minimizes the risk of forgetting prior learning, as shown by only a 3\% drop in performance metrics (see Table~\ref{tab:stages}). Meanwhile, the sharpness and level of detail in the depth maps improve significantly. All other hyperparameters, including image size, batch size, and learning rate, remain the same as in stage 2. Note that a similar strategy is also employed in the concurrent work by Bochkovskii et al.~\cite{DepthPro}.

\begingroup
\setlength{\tabcolsep}{5.5pt}
\begin{table*}
  \centering
  \small
  \caption{Comparison with current state-of-the-art methods. \colorbox{red!40}{Score} means best results, \colorbox{orange!40}{score} means second best results, and \colorbox{yellow!40}{score} means third best.}
  \vspace{-0.1in}
  \begin{tabular}{cc|ccccccccc}
    \toprule
      & Metric & Midas V3.1 & DA V1 Small & DA V2 Small & DA V1 Large & DA V2 Large & Depth Pro & Ours Stage 3\\
    \midrule
    NYUv2 & AbsRel\(\downarrow\)  & 0.0307&0.0357&0.035& \cellcolor{orange!40}0.0294& \cellcolor{orange!40}0.0294& \cellcolor{yellow!40}0.0295& \cellcolor{red!40}0.029\\ 
    TUM & AbsRel\(\downarrow\)  & 0.0439&	0.0421&0.071&\cellcolor{red!40}0.0322&0.06& \cellcolor{yellow!40}0.036&	 \cellcolor{orange!40}0.0345\\
    Sintel & AbsRel\(\downarrow\) &\cellcolor{orange!40}0.1867&0.1983&0.217&\cellcolor{red!40}0.1706&0.2117&	0.2131&	\cellcolor{yellow!40}0.1868\\
    KITTI & AbsRel\(\downarrow\)&0.0166&0.0133&0.0127&0.0131&\cellcolor{yellow!40}0.0117&\cellcolor{orange!40}0.0101&\cellcolor{red!40}0.0092\\
    ETH3D & AbsRel\(\downarrow\)  &0.0722&	0.0738&	0.0767&	\cellcolor{orange!40}0.0646&\cellcolor{yellow!40}	0.066&\cellcolor{red!40}0.0608&0.0729\\
    \midrule
    NYUv2& 100\(\cdot\)(1 - \(\delta_1\))\(\downarrow\)  & \cellcolor{orange!40}1.8887&2.8135&2.6976&\cellcolor{yellow!40}1.9683&2.1308&2.1209&\cellcolor{red!40}1.789\\
    TUM&100\(\cdot\)(1 - \(\delta_1\))\(\downarrow\)  &6.467&5.4212&7.2777&\cellcolor{red!40}3.4231&5.9623&\cellcolor{yellow!40}5.1622&\cellcolor{orange!40}4.0727\\
    Sintel&100\(\cdot\)(1 - \(\delta_1\))\(\downarrow\)  &37.7252&\cellcolor{yellow!40}33.0803&34.4476&\cellcolor{red!40}29.066&33.1902&34.0763&\cellcolor{orange!40}30.1757\\
    KITTI&100\(\cdot\)(1 - \(\delta_1\))\(\downarrow\)  &4.9082&6.1238&5.5026&5.6687&\cellcolor{yellow!40}4.5226&\cellcolor{orange!40}3.5467&\cellcolor{red!40}3.2374\\
    ETH3D&100\(\cdot\)(1 - \(\delta_1\))\(\downarrow\)  &20.7798&20.8624&20.9041&\cellcolor{red!40}19.4561&\cellcolor{orange!40}19.5112&20.2908&\cellcolor{yellow!40}19.9249\\
    \midrule
    DIW & WHDR\(\downarrow\) & 0.1112 & \cellcolor{red!40}0.1035& 0.1074 & \cellcolor{yellow!40}0.1051 & 0.1114 & 0.1288 &\cellcolor{orange!40}0.1037 \\    
    \midrule
     & Avg. rank & 
     4.6 & 4.9 & 5.9 & \cellcolor{orange!40}2.4 & 4 &  \cellcolor{yellow!40}3.6 & \cellcolor{red!40}2.1 \\
  \end{tabular}
  \label{tab:final_comparison}
  \vspace{-0.1in}
\end{table*}
\endgroup

\begin{figure*}
  \centering
  \includegraphics[width=0.95\textwidth]{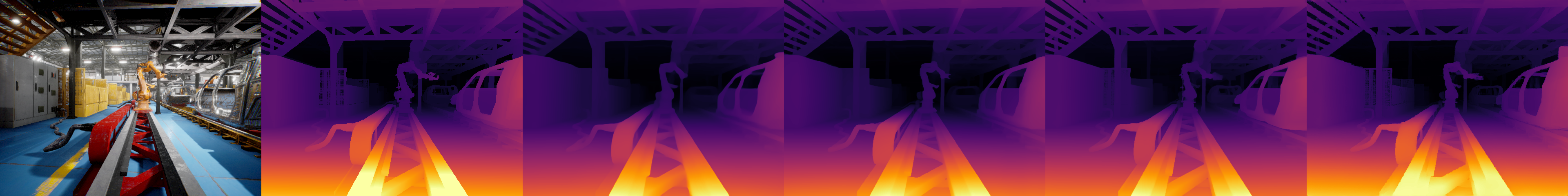}
  \includegraphics[width=0.95\textwidth]{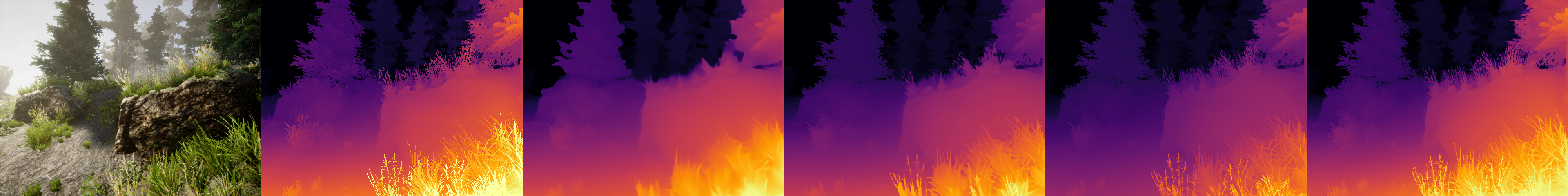}
  \includegraphics[width=0.95\textwidth]{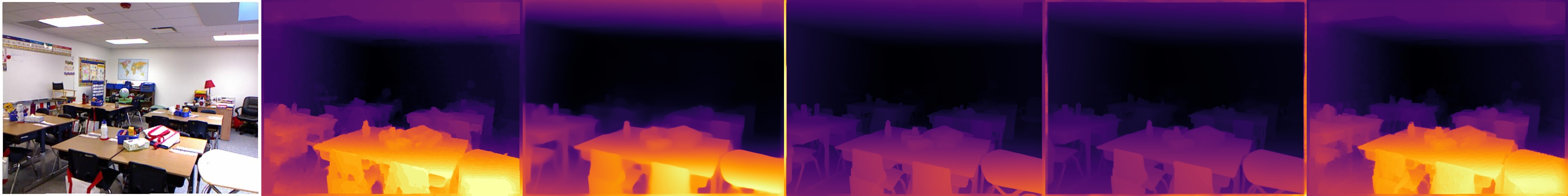}
  \includegraphics[width=0.95\textwidth]{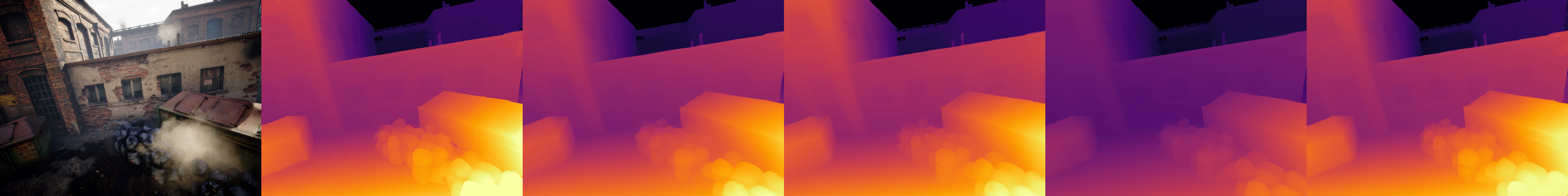}
  \includegraphics[width=0.95\textwidth]{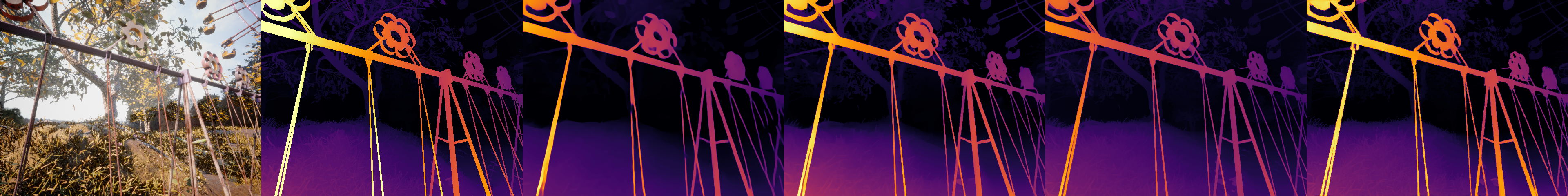}
  \includegraphics[width=0.95\textwidth]{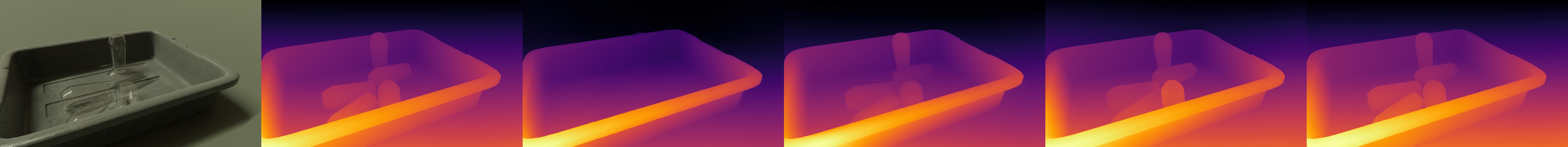}
  \includegraphics[width=0.95\textwidth]{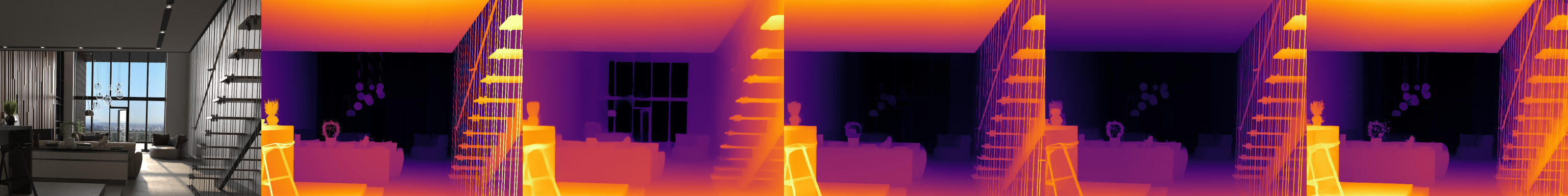}  
  \vspace{-0.1in}
  \caption{Comparison of depth output. From left to right: original image, ground truth, Depth Anything V1 Large, Depth Anything V2 Large, Depth Pro, EfficientDepth (Stage 3)}
  \label{fig:model_comparison}
  \vspace{-0.15in}
\end{figure*}

\subsubsection{Details}

For different stages of training, we adjusted the hyperparameters accordingly. We used Adam \cite{Adam} as our primary optimizer, which outperformed traditional stochastic gradient descent for this task. A cosine annealing learning rate scheduler with one epoch warmup was employed.

In terms of augmentations, we resized images by their longest side with padding, preserving the aspect ratio through the padding, which was excluded from training using a mask. We applied basic affine transformations, such as random translation, scaling, and rotation, and normalized the images. Although we experimented with other color and geometric augmentations, they did not yield significant improvements. We used Albumentations \cite{Albumentations} for CPU-based and Kornia \cite{Kornia} for GPU-based augmentations.

Most of the training was performed on 10xA40 GPUs. To reduce training time, increase stability, and accommodate larger batch sizes, we used PyTorch’s channel-last memory format, BF16 precision, and model compilation features available in newer PyTorch versions. We found BF16 to be especially beneficial, as training in FP16 mode frequently resulted in NaNs.

\begingroup
\setlength{\tabcolsep}{5.5pt}
\begin{table*}[t]
  \centering
  \small
  \caption{Speed comparison against the state-of-the-art methods. We are reporting the timing for EfficientDepth Stage 3, but the timing of our other models are similar. \colorbox{red!40}{Score} means best results, \colorbox{orange!40}{score} means second best results, and \colorbox{yellow!40}{score} means third best.}
  \vspace{-0.1in}
  \begin{tabular}{cc|ccccccc}
    \toprule
   GPU  & Metric & Midas V3.1 & DA V1 small & DA V2 small &  DA V1 large & DA V2 large & Depth Pro & EfficientDepth\\ 
    \midrule
    A40 &Avg. speed, s\(\downarrow\) & 0.46&\cellcolor{red!40}0.048&\cellcolor{red!40}0.048&0.18&\cellcolor{yellow!40}0.16&1.19&\cellcolor{orange!40}0.055\\
  \end{tabular}
  \label{tab:speed}
  \vspace{-0.1in}
\end{table*}

\begin{table}[t]
  \centering
  \caption{Ablation studies for different improvements we employed. \colorbox{red!40}{Score} means best results, \colorbox{orange!40}{score} means second best results, and \colorbox{yellow!40}{score} means third best.}
  \vspace{-0.1in}
  \begin{tabular}{c|c}
    \toprule
      & Validation \(\delta_1\)\(\downarrow\) \\ 
    \midrule
     Baseline & \colorbox{yellow!40}{0.432} \\
     + Bimodal Density Head & \colorbox{orange!40}{0.426} \\
     + $\mathcal{L}_{\text{LPIPS}}$ & \colorbox{red!40}{0.366} \\
  \end{tabular}

  \label{tab:ablations}
  \vspace{-0.1in}
\end{table}
\endgroup

\subsection{Objectives}
\label{ssec:objectives}

To ensure the network is able to learn highly detailed depth maps, we use the typical affine-invariant and edge loss and additionally propose a loss based on LPIPIS~\cite{LPIPS}. Overall, our objective consisting of three terms is defined as:

\vspace{-0.1in}
\begin{equation}
\mathcal{L} = \alpha_l\mathcal{L}_l 
 + \alpha_{edge}\mathcal{L}_{edge} + \alpha_{\mathrm{LPIPS}}\mathcal{L}_{\mathrm{LPIPS}}.
\end{equation}
\vspace{-0.1in}


Here, the first term is the scale and shift invariant loss introduced in \cite{Midas}:

\vspace{-0.1in}
\begin{equation}
    \mathcal{L}_l = \frac{1}{HW}
\sum_{i, j=1}^{H,W} \rho(d_{i,j}^\ast, d_{i,j}),
\end{equation}
\vspace{-0.1in}

\noindent where $d^\ast$ and  $d$ are the predicted and the ground truth depth maps, respectively, and $W$ and $H$ refer to the width and height of the image. Moreover, $\rho$ is the affine-invariant mean absolute error defined as:

\vspace{-0.1in}
\begin{equation}
    \rho(d_{i,j}^\ast, d_{i,j}) = \vert \hat{d_{i,j}^\ast} - \hat{d_{i,j}} \vert,
\end{equation}
\vspace{-0.1in}

\noindent where $\hat{d_{i,j}^\ast}$ and $\hat{d_{i,j}}$ are the normalized predicted and ground truth depth maps. Specifically, $\hat{d_{i,j}}$ is obtained as ($\hat{d_{i,j}^\ast}$ is computed in a similar manner):

\vspace{-0.1in}
\begin{equation}
    \hat{d_{i,j}} = \frac{d_{i,j} - t(d)}{s(d)},
\end{equation}
\vspace{-0.1in}

\noindent where
\vspace{-0.1in}
\begin{equation}
    t(d) =\mathrm{median}(d), \quad s(d) = \frac{1}{HW}\sum_{i, j=1}^{H,W} |d_{i,j}- t(d)|.
\end{equation}
\vspace{-0.1in}

The second term is the edge loss defined as:
\vspace{-0.1in}
\begin{equation}
    \mathcal{L}_{edge} = \sqrt{\frac{1}{HW}\sum_{i,j=1}^{H,W} (\hat{d_{i,j}^\ast} - \hat{d_{i,j}})^2},
\end{equation}
\vspace{-0.1in}

\noindent where $\hat{d}$ is the gradient of $d$ obtained by convolving depth with the standard Laplacian kernel 
\vspace{-0.05in}
\begin{equation}
    \left(
\begin{array}{rrr}
-1 & -1 & -1\\
-1 & 8 & -1\\
-1 & -1 & -1
\end{array}
\right).
\end{equation}
\vspace{-0.05in}

The third term is our proposed loss based on LPIPS\cite{LPIPS} and is defined as
\begin{equation}
    \mathcal{L}_{\LPIPS} = \LPIPS(\hat{d^\ast}, \hat{d})
\end{equation}

\noindent where $\hat{d^\ast}$ and $\hat{d}$ are the predicted and ground truth depth maps that are mapped to the interval $[-1, 1]$ through min-max normalization. Specifically, $\hat{d}$ is obtained as ($\hat{d^\ast}$ is computed similarly):

\vspace{-0.05in}
\begin{equation}
    \hat{d} = 2 \frac{d - \min(d)}{\max(d) - \min(d)} - 1.
\end{equation}
\vspace{-0.05in}

Finally, $\alpha_l$, $\alpha_{edge}$, and $\alpha_{\mathrm{LPIPS}}$, define the weight of the first, second, and third terms and we set them to $\alpha_l = 0.4$, $\alpha_{edge} = 0.2$, and $\alpha_{\mathrm{LPIPS}} = 0.4$.






\begin{figure*}
  \centering
  \includegraphics[width=\linewidth]{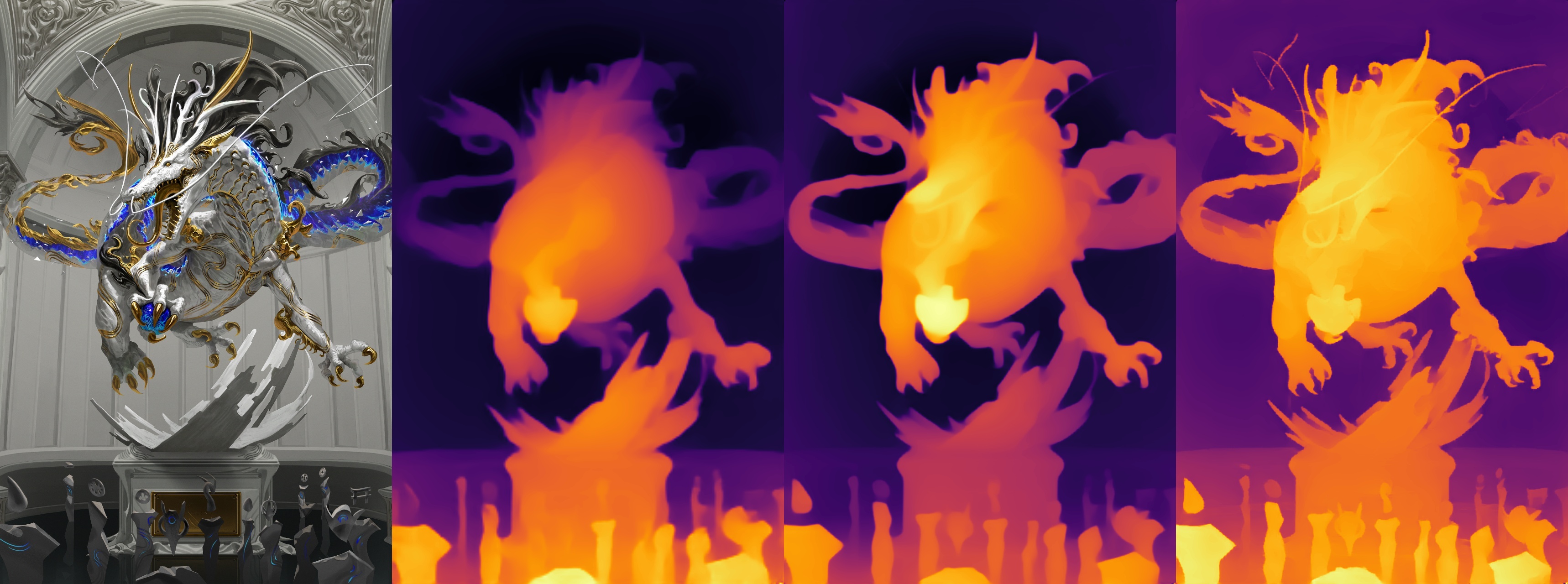}
  \includegraphics[width=\linewidth]{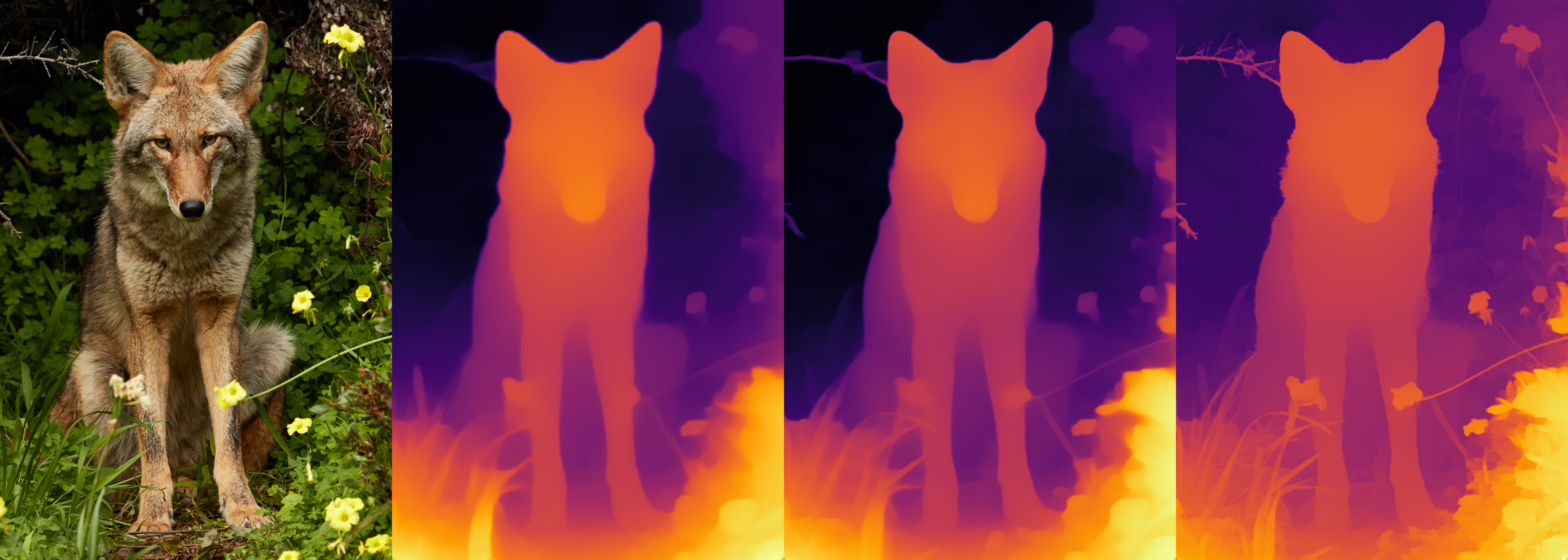}
  \hfill
  \vspace{-0.2in}
  \caption{Comparison of depth output on each stage of the training. Stage 1 (2nd column), Stage 2 (3rd column) and Stage 3 (4th column).}
  \label{fig:stages}
  \vspace{-0.1in}
\end{figure*}

\begingroup
\setlength{\tabcolsep}{3pt}
\begin{table}
\small
  \centering
    
  \caption{Results for every stage of our multi-stage training approach. Stage 3+ indicates our results after applying our \boost{}. Boosting is applied on the datasets with at least one image side more than 960. \colorbox{red!40}{Score} means best results, \colorbox{orange!40}{score} means second best, and \colorbox{yellow!40}{score} is third best.}
  \vspace{-0.1in}
  \begin{tabular}{cc|cccc}
    \toprule
     & Metric & Stage 1 & Stage 2 & Stage 3 & Stage 3+\\
    \midrule
  NYUv2 & AbsRel\(\downarrow\) & \cellcolor{yellow!40}0.0309	&\cellcolor{red!40}0.0287&\cellcolor{orange!40}0.029& --- \\
  TUM & AbsRel\(\downarrow\) & \cellcolor{yellow!40}0.0382	&\cellcolor{orange!40}0.0362	&\cellcolor{red!40}0.0345& ---\\
  Sintel & AbsRel\(\downarrow\) & 0.194&\cellcolor{red!40}0.1731&\cellcolor{yellow!40}0.1868&\cellcolor{orange!40}0.1752\\
  KITTI & AbsRel\(\downarrow\) & 0.013&\cellcolor{yellow!40}0.0096&\cellcolor{orange!40}0.0092&	\cellcolor{red!40}0.0085\\
  ETH3D & AbsRel \(\downarrow\) &\cellcolor{yellow!40}0.0715&\cellcolor{red!40}0.0665&0.0729&\cellcolor{orange!40}0.0692\\
  \midrule
  NYUv2 & 100\(\cdot\)(1 - \(\delta_1\))\(\downarrow\) & \cellcolor{yellow!40}2.0509&\cellcolor{red!40}1.7455&\cellcolor{orange!40}1.789& --- \\
  TUM & 100\(\cdot\)(1 - \(\delta_1\))\(\downarrow\) & \cellcolor{yellow!40}5.3422&\cellcolor{orange!40}4.2432&	\cellcolor{red!40}4.0727& ---\\
  Sintel & 100\(\cdot\)(1 - \(\delta_1\))\(\downarrow\) & 32.2968&\cellcolor{orange!40}29.4054&\cellcolor{yellow!40}30.1757&\cellcolor{red!40}29.0867\\
  KITTI & 100\(\cdot\)(1 - \(\delta_1\))\(\downarrow\) & 7.0097&\cellcolor{yellow!40}3.6108&\cellcolor{orange!40}3.2374&\cellcolor{red!40}2.7321\\
  ETH3D & 100\(\cdot\)(1 - \(\delta_1\)) \(\downarrow\) &20.6073&\cellcolor{orange!40}19.8421&\cellcolor{yellow!40}19.9249&\cellcolor{red!40}19.6268\\
    \midrule
    DIW & WHDR \(\downarrow\) & \cellcolor{red!40}0.0985& \cellcolor{orange!40}0.0987& \cellcolor{yellow!40}0.1037& ---\\
  \end{tabular}

  \label{tab:stages}
  \vspace{-0.1in}
\end{table}
\endgroup

\begin{figure*}
  \centering
  \includegraphics[width=\textwidth]{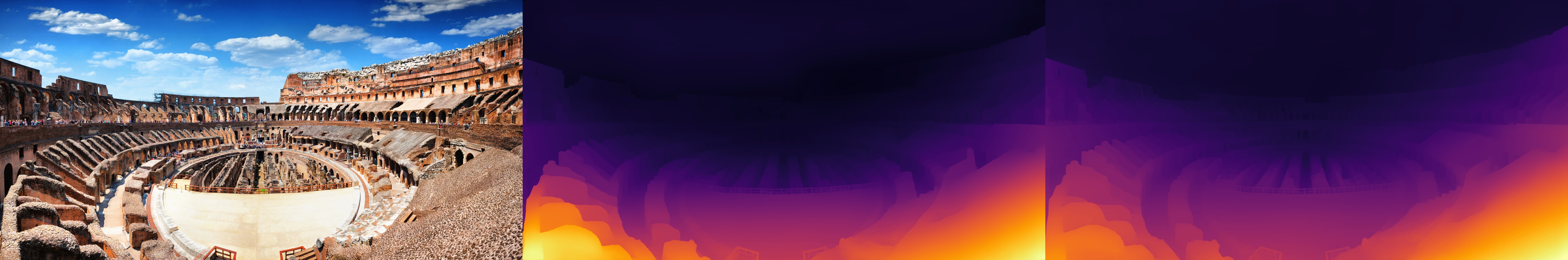}  
  \includegraphics[width=\textwidth]{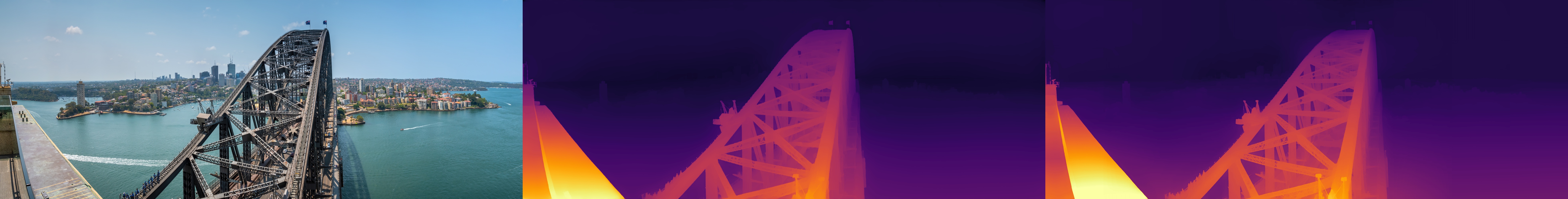}
  \includegraphics[width=\textwidth]{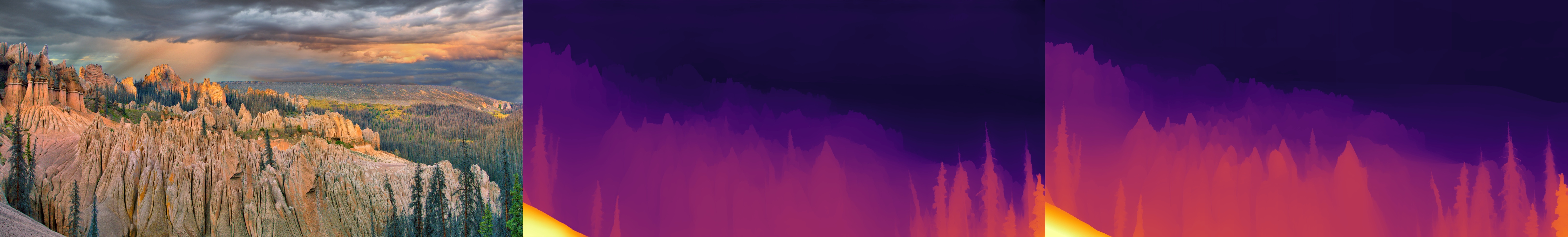}  

  \hfill
  \vspace{-0.2in}
  \caption{Comparison of our Stage 3 model without downsampling of image (2nd column) and Stage 3 + \boost  (3rd column). The difference is best observed in close-up.}
  \label{fig:boosting}
  \vspace{-0.2in}
\end{figure*}
\section{Results}

\subsection{Evaluation Metrics}
For benchmarking, we use Midas’s \cite{Midas} evaluation scripts along with widely adopted datasets for zero-shot depth estimation evaluation. For all datasets except DIW, we used 100\(\cdot\)(1 - \(\delta_1\)) and AbsRel metrics. For the DIW dataset, which is human-labeled, we used the WHDR metric, specifically designed to evaluate depth estimation on such datasets.

\subsection{Qualitative and Quantitative Comparisons}
We compare our model with Midas V3.1~\cite{Midas}, Depth Anything V1~\cite{DA1}, Depth Anything V2~\cite{DA2}, and the concurrent work, Depth Pro~\cite{DepthPro}. The results, presented in Table~\ref{tab:final_comparison}, were generated using our benchmarking script, so values may differ from those reported by the original authors. Still, we used all the preprocessing parameters the authors mentioned in their work. 

As is often the case with depth benchmarks, the performances vary across different datasets. For example, our model performs best on the NYUv2 and KITTI datasets and ranks second on the TUM and DIW, while Depth Anything V1 achieves top results on the TUM and Sintel datasets. These results demonstrate that using Depth Anything V1 as the pseudo-labeling model does not limit our model’s potential quality; in fact, we are able to surpass the performance of the teacher model.


Moreover, we present visual comparisons with other approaches in Fig.~\ref{fig:model_comparison}. Overall, EfficientDepth produces highly detailed and geometrically consistent depth maps. In terms of detail, our results are comparable (e.g., stairs in last row) or superior (e.g., grass blades in second row) to all other approaches. This performance is achieved while being significantly more efficient than the other high-performing methods, as discussed next.


\subsection{Speed Comparisons}
For the speed benchmark in Table~\ref{tab:speed}, we created a dataset of 500 images with varying resolutions. The benchmark measures the average complete run time of each model, including all preprocessing and postprocessing steps (but without image loading) to ensure a fair comparison. All models were tested on a standard AWS instance with a single A40 GPU. All models were run in exactly the same way as for quality comparison in Table~\ref{tab:final_comparison}. As seen, our network is significantly faster than the high-performing methods in Table~\ref{tab:final_comparison}. Our efficiency is comparable to the small variant of Depth Anything V1 and V2, however, as shown in Table~\ref{tab:final_comparison}, our depth maps are more accurate.




\subsection{Ablation studies}

\paragraph{Effect of Various Components} Here, we discuss the impact of our major improvements, specifically the Bimodal Density Head and LPIPS loss. Table~\ref{tab:ablations} shows the results of each iteration on our Stage 1 model. We used validation \(\delta_1\) as the primary metric, representing the percentage of depth pixels with a relative error exceeding 25\%, so a lower score indicates better performance. The Bimodal Density Head contributed approximately 2\% relative improvement, while the addition of LPIPS loss provided a significant boost, improving results by an additional 15\%.

\paragraph{Multi-Stage Training}
Results from the different training stages are shown in Fig.~\ref{fig:stages}, while Table~\ref{tab:stages} provides a detailed comparison of benchmark results for each stage. As seen, Stage 1 performs reasonably well, but Stage 2 offers a substantial improvement, boosting performance by over 25\% on average. Stage 2 achieves the best metrics overall, having been trained on the largest dataset. However, the depth predictions at this stage still lack sharpness, as illustrated in Fig.~\ref{fig:stages}. This is because the network at this stage is trained on a combination of real datasets and datasets with pseudo-labels, which inherently limit the level of detail due to label inaccuracies. The final stage introduces sharpness and enhances details, with a minor average quality reduction of around 3\%.  The increased sharpness is achieved because of training the network on a subset of synthetic data with accurate labels. However, since the training data at this stage is limited, the network forgets some of the priors learned from the full training set, leading to a slight drop in quantitative metrics. The increased detail is significantly beneficial for tasks like view synthesis and 3D reconstruction, where capturing depth discontinuities at object edges has a greater impact on perceptual quality than small metric gains.


\paragraph{\boost{}}
Results of our \boost{} approach are shown both numerically and visually in Table~\ref{tab:stages} and Fig.~\ref{fig:boosting}, respectively. As seen, \boost{} improves the accuracy and level of detail in the depth maps, at the cost of additional computation (see supplementary materials). Note that \boost{} can be applied to models from any training stage, allowing flexibility based on application needs such as speed, geometric accuracy, or level of detail.


\paragraph{Handling Reflective Surfaces}
Our method is specifically designed for applications like view synthesis. In such cases, accurately estimating the depth of reflected objects, rather than the reflective surface itself, is more critical as it ensures correct reprojection of these objects in novel views. A few example results with comparisons against existing techniques are shown in the supplementary materials.


\section{Conclusion}

In conclusion, EfficientDepth offers a robust solution to monocular depth estimation, delivering both high geometric consistency and fine detail essential for real-world applications such as 3D reconstruction and view synthesis. This is achieved through a carefully designed lightweight network architecture trained on a diverse corpus of labeled and pseudo-labeled images. Specifically, we generate pseudo labels using an existing depth estimation method, enhancing its predictions with a simple yet effective boosting strategy. Additionally, we implement a multi-stage training approach and introduce a perceptual loss function to improve the network’s ability to produce detailed depth maps. Extensive experiments show that our model achieves comparable or better results across several benchmarks while maintaining greater efficiency.



{
    \small
    \bibliographystyle{ieeenat_fullname}
    \bibliography{main}
}

\clearpage
\maketitlesupplementary

%
\begin{figure*}
  \centering
  \includegraphics[width=\textwidth]{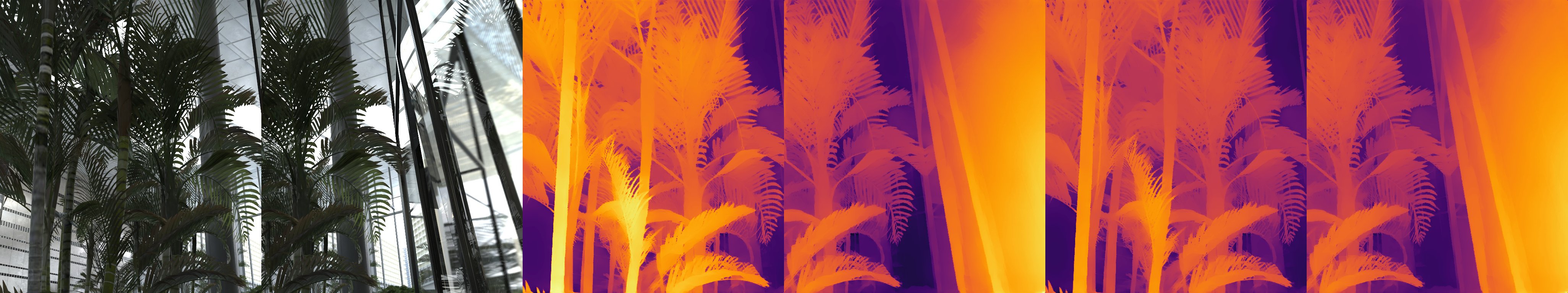}
  \includegraphics[width=\textwidth]{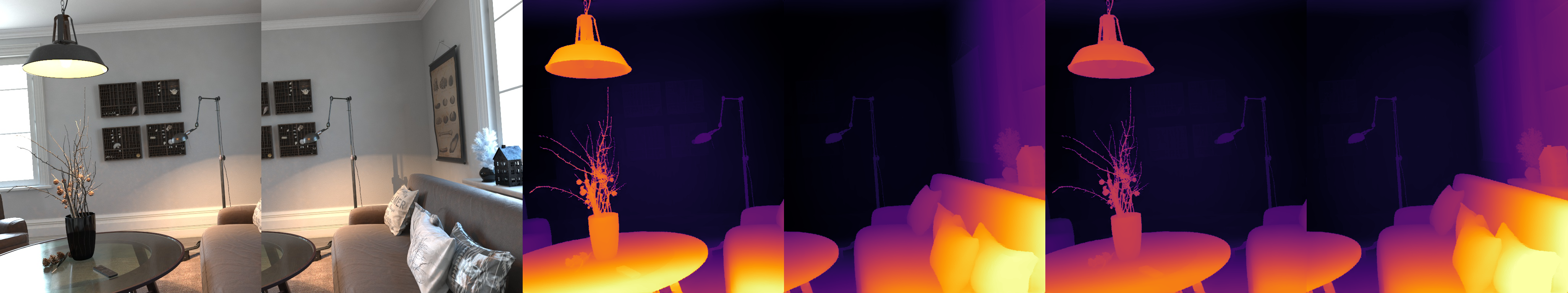}
  \vspace{-0.2in}
  \caption{An illustration of our boosting algorithm. From left to right: overlapping patches, Stage 3 applied to the patches separately, the same patches after applying the scale and offset.}
  \label{fig:overlapping}
  \vspace{-0.15in}
\end{figure*}

\section{LPIPS rationale}
Although LPIPS was originally designed for RGB images, the discontinuities and edges in depth maps are similar to those in images. As such, it effectively captures structural and detail similarities in depth maps, and our experiments demonstrate that using LPIPS leads to substantial improvements in detail preservation.

\section{Model architecture}
The features extracted by MiT-B5 are utilized similarly to those from other encoders (e.g., ResNet), with multi-scale features fed into the decoder at each corresponding layer, ensuring seamless integration with the UNet architecture. Unlike the original SegFormer paper \cite{SegFormer}, which employs transformers for both encoding and decoding, and approaches like \cite{Umixformer}, our method uses MiT-B5 solely for feature extraction during the encoding stage, while the decoder remains fully convolutional. This design enables strong performance without a significant computational overhead and allows for dynamic input sizes. The implementation can be found in \cite{SMP}.


\section{SimpleBoost}
In Fig.~\ref{fig:overlapping}, we illustrate the internal stages of our boosting algorithm. First, we divide the image into overlapping patches (1st column) and apply the depth estimation model to each patch separately (2nd column). Since the scale of each patch is inconsistent, we perform a linear scaling and shifting of the depth values for each patch (3rd column). Finally, we blend the overlapping regions to produce the final depth map (not shown).


\paragraph{Comparison with Boosting Monocular Depth \cite{Boosting}}

\begin{figure*}
  \centering
  \includegraphics[width=0.95\textwidth]{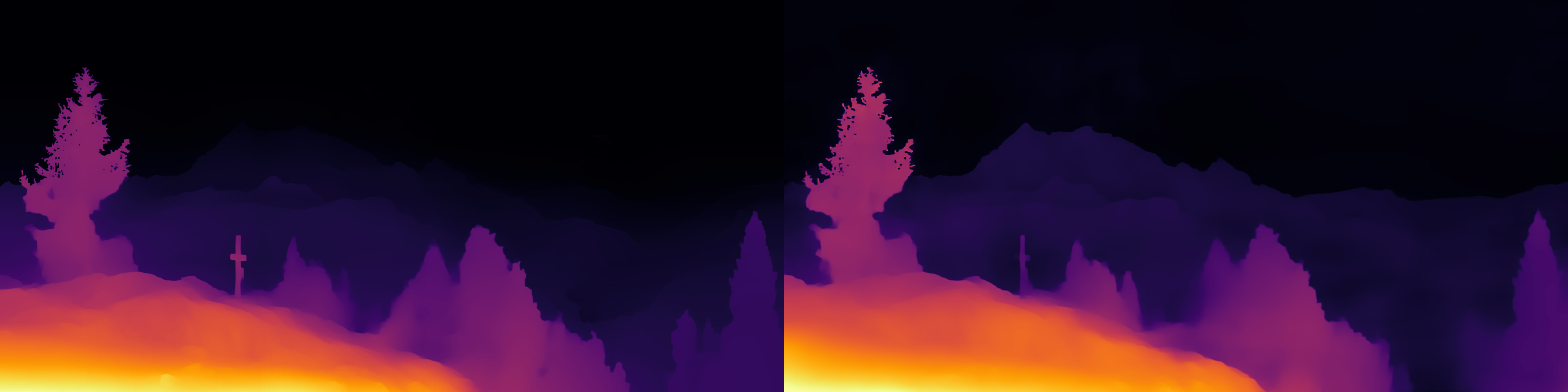}
  \hfill
  \vspace{-0.1in}
  \caption{Cross and the full picture. Ours (left), Boosting Monocular Depth (right)}
  \label{fig:boosting-comparison}
  \vspace{-0.15in}
\end{figure*}

We provide a visual comparison with the boosting approach of Miangoleh et al in Fig.\ref{fig:boosting-comparison}. While their method produces reasonable results, our SimpleBoost more accurately reconstructs fine structures, such as the cross. Since their model is trained using MiDaS V2, we apply both their method and ours on top of MiDaS V2 to ensure a fair comparison. In terms of inference time, our SimpleBoost adds an average overhead of 0.77s, whereas the method by Miangoleh et al. introduces a larger overhead of 1.86s. Compared to processing the full-resolution image directly, SimpleBoost processes smaller patches sequentially, which reduces peak memory usage—making it especially suitable for high-resolution inputs on memory-constrained devices.

\section{Handling Reflective Surfaces}
\begin{figure*}
  \centering
  \includegraphics[width=\textwidth]{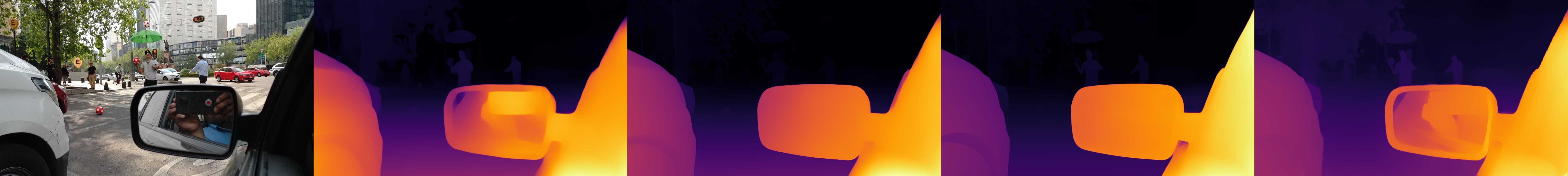}
  \includegraphics[width=\textwidth]{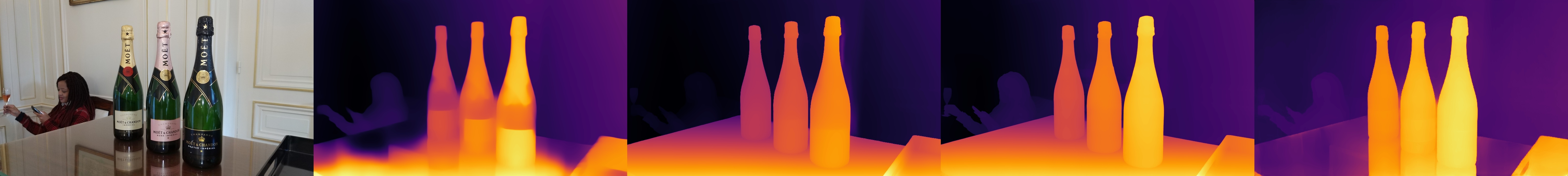}
  \hfill
  \vspace{-0.1in}
  \caption{Comparison of depth output in the wild (no ground truth). From left to right: original image, Depth Anything V1 Large, Depth Anything V2 Large, Depth Pro, EfficientDepth (Stage 3)}
  \label{fig:reflections}
  \vspace{-0.15in}
\end{figure*}

Our method is specifically designed for applications like view synthesis. In such cases, accurately estimating the depth of reflected objects, rather than the reflective surface itself, is more critical as it ensures correct reprojection of these objects in novel views. As seen in Fig.~\ref{fig:reflections}, Depth Anything V2 Large and Depth Pro estimate the depth of the reflective surface. Depth Anything V1 Large predicts the depth of the reflected objects, similar to ours, but is less detailed (car's mirror image, mirror table).

\end{document}